\documentclass[runningheads]{llncs}

\usepackage[T1]{fontenc}

\usepackage{graphicx}
\usepackage{comment}
\usepackage{amsmath,amssymb}
\usepackage{color}
\usepackage{url}
\usepackage{hyperref}
\usepackage[nocompress]{cite}
\usepackage{graphicx}
\usepackage{amsmath}
\usepackage{amssymb}
\usepackage{xcolor}
\usepackage{bm}
\usepackage{booktabs}
\usepackage{algorithm}
\usepackage{algpseudocode}
\usepackage{dsfont}
\usepackage{cleveref}
\usepackage{adjustbox}
\usepackage{multirow,multicol}
\usepackage{subcaption}
\usepackage{tikz}
\usepackage{pgfplots}
\pgfplotsset{compat=1.18} 
\DeclareMathOperator*{\argmax}{arg\,max}
\DeclareMathOperator*{\argmin}{arg\,min}

\newif\ifreview
\reviewfalse

\ifreview
	\usepackage{lineno}

	\linenumbers
\fi

\begin{document}


\def\SubNumber{047}


\title{Structured Universal Adversarial Attacks on Object Detection for Video Sequences}
\titlerunning{Structured Universal Adversarial Attacks}

\ifreview
\else

	\author{Sven Jacob\inst{1,2} \and
	Weijia Shao\inst{1} \and
	Gjergji Kasneci \inst{2,3}}
	
	\authorrunning{Jacob et al.}
	
	\institute{Federal Institute for Occupational Safety and Health (BAuA), Dresden, Germany
	\email{\{jacob.sven,shao.weijia\}@baua.bund.de}\\
	 \and School of Computation, Information and Technology,
Technical University of Munich, Munich, Germany
    \and School of Social Sciences and Technology,
Technical University of Munich, Munich, Germany}
\fi

\maketitle              

\begin{abstract}
Video-based object detection plays a vital role in safety-critical applications. While deep learning-based object detectors have achieved impressive performance, they remain vulnerable to adversarial attacks, particularly those involving universal perturbations. In this work, we propose a minimally distorted universal adversarial attack tailored for video object detection, which leverages nuclear norm regularization to promote structured perturbations concentrated in the background. To optimize this formulation efficiently, we employ an adaptive, optimistic exponentiated gradient method that enhances both scalability and convergence. Our results demonstrate that the proposed attack outperforms both low-rank projected gradient descent and Frank-Wolfe-based attacks in effectiveness while maintaining high stealthiness. All code and data are publicly available at \url{https://github.com/jsve96/AO-Exp-Attack}.

\keywords{Adversarial Attacks  \and Object Detection \and AI Safety \and Robustness}
\end{abstract}
\section{Introduction}
Video-based object detection plays an increasingly important role in safety monitoring systems for machine and occupational environments, enabling the localization of human workers, tools, and obstacles to identify potential hazards before they escalate into accidents \cite{alkaabi2023methodology,cocca2016video,malburg2021object}. As a core task in computer vision, object detection involves identifying and localizing semantic objects within images or videos. Recent advances in deep learning have significantly improved object detection performance, enabling its deployment in a range of safety-critical domains, ranging from autonomous driving \cite{feng2020deep_driving1,Chen_2016_CVPR_driving2}, real-time surveillance \cite{VOT_Example_Surveillance2,VOT_Example_Surveillance}, and industrial applications \cite{CV_Industrial,CV_Industrial_2}. In these contexts, object detection not only contributes to operational efficiency but also serves as a first line of defense in preventing unsafe interactions between humans and machines.

\begin{figure}
    \centering
    \includegraphics[width=0.99\linewidth]{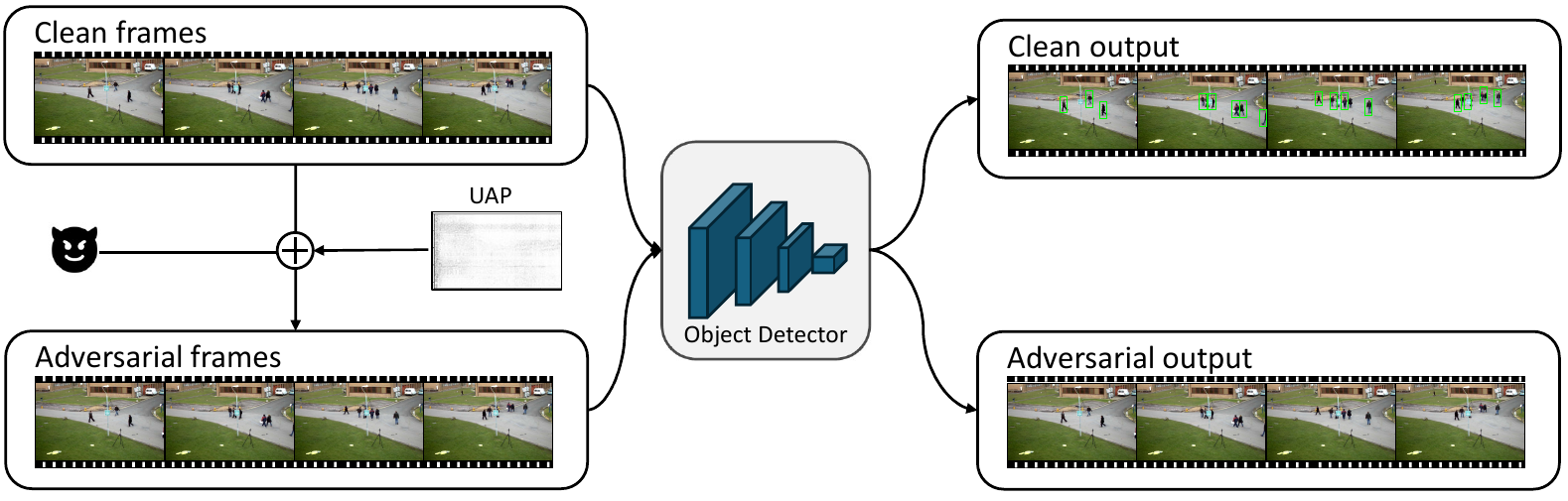}
    \caption{Shows conceptual framework of universal adversarial attack on object detector. A carefully crafted Universal Adversarial Perturbation (UAP) suppresses all bounding boxes after applied on clean frames.}
    \label{fig:FRAMEWORK}
\end{figure}
Most state-of-the-art object detection methods rely on Deep Learning (DL) techniques \cite{8627998}. Despite their substantial advancements over the past decade, DL models are vulnerable to adversarial attacks (AT) \cite{goodfellow2014explaining,szegedy2013intriguing}, which craft perturbations to the inputs to mislead the model into making incorrect predictions. While research on adversarial attacks in image classification has been extensively studied \cite{ding2020adversarial}, such attacks on object detection systems, especially in the context of video data, have received considerably less attention \cite{nguyen2025survey}. At first glance, adversarial attacks on video object detection may appear straightforward, as they seem to require applying existing techniques for attacking static object detectors to each frame of the video clip \cite{thys2019fooling,WANG2020102634,9428443}. In \cite{li2021universal}, the authors have empirically proved the existence of universal adversarial perturbations against all frames, which cause object detectors to fail on most of the frames \cite{li2021universal,pmlr-v101-wu19a}. Effective universal attacks pose a more significant threat as they are transferable
across frames without further accessing the target model, and are more convenient to be applied in the real physical world \cite{nguyen2025survey,li2021universal}.

Although prior studies have aimed to generate adversarial examples posing greater threats, they have predominantly focused on perturbations bounded by $\ell_2$ and $\ell_\infty$ norms \cite{nguyen2025survey}. In image-based attacks, $\ell_1$-bounded perturbations can be especially threatening due to their ability to introduce sparse yet imperceptible changes \cite{croce2021mind}. However, in video-based settings, the direct application of $\ell_1$ attacks often results in visible patches on moving objects across frames. This not only reduces the sparsity but also challenges stealthiness in dynamic scenes. As adversarial defense mechanisms continue to evolve, identifying a broader range of attack strategies is essential for robust evaluation and understanding of their limitations. 

\begin{figure}
    \centering
    \includegraphics[width=0.99\columnwidth]{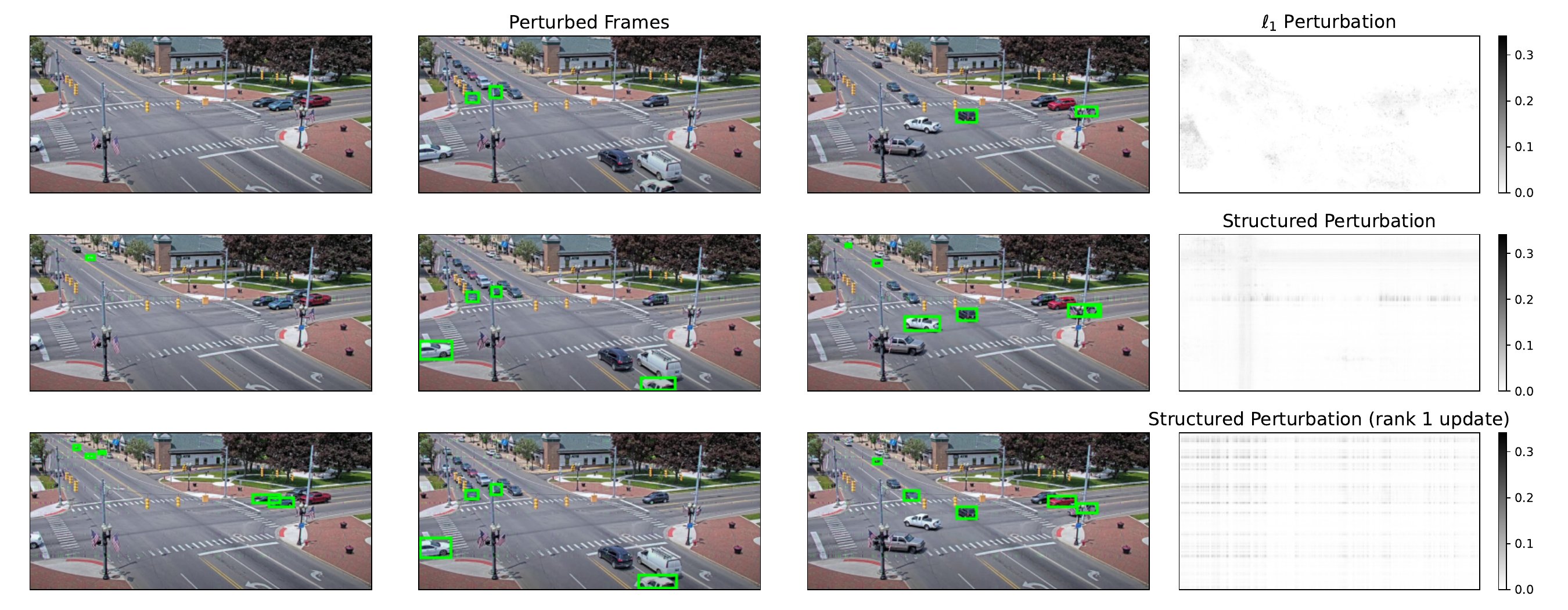}
    \caption{The $\ell_1$ attack introduces flickering noise that trails moving objects and spreads across the street in subsequent frames, whereas the nuclear norm-based attack perturbs orthogonal spatial patterns of the video frames, resulting in more structured and spatially coherent perturbations.}
    \label{fig:L1-vs-Nuc}
\end{figure}

Building on robust principal component analysis for segmentation \cite{8425659} and structured adversarial perturbation methods for image classification \cite{kazemi2023minimally}, this work introduces a novel strategy that leverages structured but non-suspicious background modifications for object vanishing attacks. The conceptual framework is illustread in \Cref{fig:FRAMEWORK}. To this end, we propose a minimally distorted attack method based on nuclear norm regularization. \Cref{fig:L1-vs-Nuc} provides a preliminary visual comparison illustrating the difference between $\ell_1$ attacks, which tend to produce sparse patches on moving objects, and our nuclear norm-based attack, which generates more structured perturbations primarily in the background. While nuclear norm regularization provides a powerful tool for promoting low-rank structure in background perturbations, it poses significant optimization challenges. To address this, we employ optimistic exponentiated gradient descent \cite{shao_optimistic_2022}, which enables efficient and scalable optimization under nuclear norm regularization. We evaluate our proposed object vanishing attack on public video datasets and video object detection models. The results demonstrate that our method effectively generates structured background perturbations that consistently remove the bounding boxes predicted by these models. Compared to existing nuclear norm-based attack approaches, our method achieves superior attack success while being significantly more computationally efficient. 

Our main contributions are summarized as follows:
\begin{itemize}
    \item We introduce a minimally distorted universal attack formulation based on nuclear norm regularization, which promotes structured perturbations of the orthogonal spatial patterns across video frames (\Cref{eq:UAT Formulation}).
    \item To efficiently solve the associated optimization problem, we adapt an adaptive optimistic exponentiated gradient descent method, enabling scalable optimization under nuclear norm constraints (\Cref{alg:cap}).
    \item We conduct comprehensive evaluations on public video datasets and a state-of-the-art video object detection model, demonstrating that our method consistently suppresses bounding boxes through subtle background changes (\Cref{fig: cd-Plot}, \Cref{fig:SVD_plot_methods}).
    \item Our method achieves superior attack success and computational efficiency compared to existing nuclear norm-based adversarial attacks (\Cref{ViT Table}).
\end{itemize}

The rest of the paper is organized as follows. \Cref{sec:related} reviews related work, and \Cref{sec:notation} introduces the notation used throughout the paper. In \Cref{sec:main}, we present our attack algorithm, which is evaluated on real-world video object detection datasets and the popular object detection model Mask-RCNN in \Cref{sec:exp}. Finally, \Cref{sec:conclusion} concludes the paper by discussing the limitations of our approach and outlining future research directions.

\section{Related Work}
\label{sec:related}
While the literature on adversarial attacks in image classification is extensive, research on adversarial attacks targeting object detection, especially in video settings, remains relatively limited. We refer to recent surveys \cite{nguyen2025survey,amirkhani2023survey} for an overview of existing techniques and challenges in this area. This work focuses on object vanishing attacks, which aim to make the model fail to detect certain objects in the input frames. Several works have explored object vanishing attacks on images by manipulating bounding box outputs using adversarial perturbations, either constrained within $\ell_2$ or $\ell_\infty$ norm balls or applied as localized patches on foreground objects \cite{pmlr-v101-wu19a,thys2019fooling,9102805,li2021universal,WANG2020102634,9428443}. 

The methods listed above can be directly applied to each frame of a video independently.
However, applying attacks separately to each frame ignores the temporal coherence of video data, often resulting in flickering perturbations across frames that reduce stealth. Universal adversarial perturbations \cite{moosavi2017universal}, which are input-agnostic and applied uniformly across inputs, offer a more natural fit for object detection in video settings \cite{li2021universal,pmlr-v101-wu19a}. Yet, the existing universal attacks \cite{li2021universal,pmlr-v101-wu19a} apply noise uniformly over the entire image without leveraging the spatial structure unique to videos.

The investigation of low-rank structures for high-dimensional problems has a long history following the manifold hypothesis, which states that high-dimensional data tend to live near a lower-dimensional manifold \cite{ManifoldHypo}. This assumption led to many dimensionality reduction methods in general but was also utilized in crafting adversarial perturbations using Autoencoder \cite{latent_space_advexp,feng2019learning_autoenc_advexp}, PCA \cite{kim2021channel_PCA_ADV_ATTACK,kravchik2021efficient_PCA_ADVAttack}, or UMAP \cite{subhash2023universal_expadv}. More recently, researchers have also focused on low-rank representation for adversarial attacks induced by nuclear norm regularization \cite{kazemi2023minimally}. Moreover, a combination of nuclear norm regularization and optimizing adversarial perturbation using projected gradient descent (PGD) was introduced in \cite{LoRaPGD}. In \cite{GSE_ATtacks}, a group-wise sparse attack is generated, which only perturbs a few semantically meaningful areas of an image.

\section{Notation}
\label{sec:notation}
Throughout the paper, we write $x\in [0,1]^{H\times W\times C}$ for an image with height $H$, width $W$, and $C$ channels. Suppose an object detection model $f:[0,1]^{H \times W \times C} \to \mathcal{S}$ that takes an input image and outputs a set $\mathcal{S}$. The minimal output set is a collection of bounding boxes and labels for each detected object within an image. The output depends on the choice of $f$, e.g., Mask R-CNN \cite{he2017mask} additionally outputs confidence scores $\xi_{i}\in(0,1]$ and masks $m_{i}\in[0,1]^{H\times W}$ where $i$ is the index of the corresponding bounding box in $\mathcal{S}$. For vector-valued inputs $x\in \mathbb{R}^{n}$, the $\ell_{p}$-norm is given by $||x||_{p}=(\sum_{i=1}^{n}|x_i|^{p})^\frac{1}{p}$ for $p\ge1$. 

Recall, for a matrix $A \in \mathbb{R}^{H \times W}$ its singular value decomposition (SVD) is $A=U\Sigma V^{T}$, where $U \in \mathbb{R}^{H \times H}$ and $V \in \mathbb{R}^{W\times W}$ are orthonormal matrices whose columns are the orthonormal basis for the column and row space of $A$, and $\Sigma \in \mathbb{R}^{H \times W}$ is a rectangular diagonal matrix storing the singular values $\sigma_{i} \ge 0$  for $i=1,\dots,r$ where $r = \min\{H,W\}$ is the rank of $A$. Define the function \[
\operatorname{diag}: \mathbb{R}^r\to\mathbb{R}^{H\times W}, \sigma\mapsto \Sigma
\text{ with } 
\Sigma_{ij}=\begin{cases}
			\sigma_i, & \text{if } i=j\\
            0 & \text{otherwise.}
		 \end{cases}
\]
The SVD decomposition can be equivalently written as $A=U\operatorname{diag}(\sigma) V^{T}$.

The Schatten p-norm of a matrix $A$ is the $\ell_{p}$-norm of the vector of its singular values $||A||_{p}=\left (\sum_{i=1}^{r} (\sigma_{i})^{p}\right )^\frac{1}{p}.$
The matrix norm for $p=2$ is also called Frobenius norm, notated as $||A||_{F}$.
The nuclear norm ($p=1$) of a matrix $A$ is the $\ell_{1}$-norm of its singular values $||A||_{*}=\sum_{i=1}^{r}\sigma_{i}.$

For $p=\infty$, we have the spectral norm of a matrix, which is the largest singular value,
$||A||_{\infty}=\max_{1\leq i\leq r} \sigma_{i}.$

\section{Adversarial Attack Formulation}
\label{sec:main}
The goal of an adversarial attack is to determine some minimal perturbation $\mathbf{\delta}$ which maximizes some loss function $\mathcal{L}:\mathcal{S} \mapsto \mathbb{R}_{+}$ of the object detector $f$, when added to $x$. A non-targeted adversarial attack can be formulated as,
\begin{equation*}
\begin{aligned}
        \min_{\mathbf{\delta} \in \mathbb{R}^{H \times W \times C}} \quad & -\mathcal{L}(f(x+\mathbf{\delta}),f(x)) + \lambda \mathcal{R}(\mathbf{\delta}) \\
        \textrm{s.t.} \quad & x+\delta \in  [0,1]^{H \times W \times C}
\end{aligned}
\end{equation*}
where $\mathcal{R}$ denotes some regularization of the perturbation and $\lambda >0$ is the regularization parameter. In the following, we propose to split the loss function into background and foreground loss, $\mathcal{L}=\mathcal{L}_{\text{fg}}+\mathcal{L}_{\text{bg}}$.
Consider the set of clean masks $\mathcal{M}=\{m_{i} |\xi_{i}>\tau; i \in [S]\}$ obtained with $f(x)$, with a confidence score above $\tau$. We combine all clean masks in $\mathcal{M}$ to obtain a unified mask $\bm{m}=\sum_{i \in \mathcal{M}}m_{i}$ for all confident detections and derive a binary mask 
\begin{equation*}
    \bm{y}_{ij} = \begin{cases}
    1 & \text{if $\bm{m}_{ij}>0$} \\
    0 & \text{otherwise}
    \end{cases}
\end{equation*}
which we use as the ground-truth segmentation for the clean image. We separate the mask into foreground pixels $\mathcal{F}$ and background pixels $\mathcal{B}$ and calculate the average cross-entropy loss between $\bm{y}$ and the predicted masks which are the output of $f(x+\delta),$
\begin{equation*}
    \mathcal{L}_{\text{fg}}=\frac{1}{|\mathcal{F}|}\sum_{i \in \mathcal{F}}\text{CE}(p_{i},y_{i}), \quad \mathcal{L}_{\text{bg}}=\frac{1}{|\mathcal{B}|}\sum_{i \in \mathcal{B}}\text{CE}(p_{i},y_{i}).
\end{equation*} 
Additionally, we seek to guide the model into less confident predictions of bounding boxes, therefore we introduce the confidence loss
\begin{equation*}
    \mathcal{L}_{\text{conf}} = \sum_{i \in [S]} \xi_{i} \cdot\mathbf{1}_{(\xi_{i}>\tau)} 
\end{equation*}
which penalizes any predictions above threshold $\tau$ - affecting the foreground. In summary, we use a combination of\[\mathcal{L}_{\textrm{total}}=\underbrace{\alpha  \mathcal{L}_{\text{fg}} + \gamma  \mathcal{L}_{\text{conf}}}_{\text{Foreground}}+\underbrace{\beta \mathcal{L}_{\text{bg}.}}_{\text{Background}}\]
Usually, regularization is introduced to ensure a sparse perturbation. For vector-valued inputs, common choices are $\ell_{p}$-norms or the Frobenius $\ell_1$-norm, which suppresses the perturbation to have few non-zero values.
The nuclear norm is related to the rank of the matrix where the minimization of it leads to sparsity in singular values, resulting in a low-rank matrix. Nuclear norm regularization is popular among image denoising \cite{gu2014weighted_imagedenoise} and low-rank matrix approximation. It has also shown promising results in domain generalization \cite{domain_gen_with_nucnorm}. 

We introduce a combination of Frobenius norm and nuclear norm as the regularizer
\begin{equation*}
    \mathcal{R}(\delta) = \lambda_{1}||\delta||_{*} + \lambda_{2}||\delta||_{F},
\end{equation*}
which balances sparsity and low-rank of the universal adversarial perturbation.
\subsection{Universal Attack}\label{Universal Attack Formnulation}
The objective of this work is to generate an adversarial perturbation that is applied uniformly across all frames of a video and degrades the performance of a target model $f$. 
Let $\{x_b|1\leq b\leq B\}$ be the set of frames in a video clip, where $B$ is the number of frames. We formalize the attack as the following regularized optimization problem
\begin{equation}\label{eq:UAT Formulation}
    \min_{\mathbf{\delta} \in \mathbb{R}^{H \times W \times C}} -\frac{1}{B}\sum_{b=1}^{B}\mathcal{L}_{\textrm{total}}(f(x_{b}+\mathbf{\delta}),f(x_{b})) +\sum_{c=1}^C (\lambda_1 ||\mathbf{\delta^c}||_{*}+\frac{\lambda_2}{2}||\mathbf{\delta^c}||_{F}^{2}).
\end{equation}
Denote the gradient information 
\[
\nabla\mathcal{G}(\delta^c)= \frac{1}{B}\sum_{b=1}^{B}\nabla_{\delta^c}\mathcal{L}_{\textrm{total}}(f(x_{b}+\delta^c),f(x_{b})),
\]
which is the loss gradient averaged over all frames concerning a perturbation channel $\delta^c$. Given the access to $\nabla\mathcal{G}(\delta^c)$, we iteratively update each perturbation channel $\delta^c$ by applying the adaptive optimistic exponentiated method (AO-Exp) proposed in \cite{shao_optimistic_2022}. The algorithm is described in Algorithm \ref{alg:cap}. 

At each iteration $t$, in addition to the perturbation $\delta_t^c$, we maintain a decision variable in the form of factors of an SVD decomposition 
\[
\delta_{t}^{c}=U_{c,t}\operatorname{diag}(z_t^c) V_{c,t}^{\top} 
\] for each channel, where $z_t^c\in \mathbb{R}_{\geq 0}^{\min\{W,H\}}$ is the vector of singular values and $U_{c,t}\in\mathbb{R}^{W\times W}$, $V_{c,t}\in\mathbb{R}^{H\times H}$ are the orthogonal bases. 
To obtain the intermediate perturbation, we apply the following procedure:\\

\noindent \textbf{(1) \textsc{Optimistic update:}}
We first perform an optimistic update using the gradient information at iterations $t$ and $t-1$ 
\begin{equation}\label{(1)OptimisitcUpdate}
    \begin{split}    
\eta^c_t\gets&\eta^c_{t-1}+t^2\lVert\nabla\mathcal{G}(\delta_t^c)-\nabla\mathcal{G}(\delta_{t-1}^c)\rVert_\infty^2\\
\bar z^c_{t,i}\gets&\log (z^c_{t,i}+1) \text{ for all } i\in\{1,\ldots,\min\{W,H\}\} \\
U_{c,t+1} \operatorname{diag}(\theta^c_t) V_{c,t+1}^{\top}\gets& \eta_{t}^{c}\cdot U_{c,t}\operatorname{diag}(\bar z^c_{t})V_{c,t}^\top +(2t+1)\nabla\mathcal{G}(\delta_t^c)-t\nabla\mathcal{G}(\delta_{t-1}^c)\\
\end{split}
\end{equation}
to obtain the orthogonal bases of iteration $t+1$.

\noindent \textbf{(2) \textsc{Singular values of decision variable:}}
Then, we find the singular values of the decision variable at iteration $t+1$ by calculating the principal branch of the Lambert function 
\begin{equation}\label{(2)Singular Values}
    z^c_{t+1,i}=\frac{\eta_{t}^{c}}{\lambda_2}W_0\left (\frac{\lambda_2}{\eta_{t}^{c}}\exp\left(\frac{\lambda_2+\max\{\theta^c_{t,i}-\lambda_1,0\}}{\eta_t}\right)\right)-1.
\end{equation}

\noindent \textbf{(3) \textsc{Construct perturbation:}}
Finally, we obtain the perturbation $\delta_{t+1}^c$ by taking the weighted average of $z^c_1,\ldots,z^c_{t+1}$. We propose to use the top $k$ values of the decision variable $z_{t}^{c}$ in the reconstruction of $\delta_{t+1}^{c}$ which further compresses the information in $\delta$, thus promoting low-rank with. Let $z_{t,1:k}^{c}=(z_{t,1}^{c},\dots,z_{t,k}^{c},0,\dots,0)$, then the low-rank perturbation is obtained with
\begin{align}\label{(3) low rank perturbation}
    \delta_{t+1}^c=\frac{2}{t(t+1)}\sum_{s=1}^{t}s \cdot U_{c,t}\operatorname{diag}(z^c_{s,1:k})V_{c,t}^\top.
\end{align}

The per-iteration complexity of the algorithms depends on the complexity of performing the SVD-decomposition and solving the principal branch of the Lambert function.

\begin{algorithm}[H]
\caption{AO-Exp Update}\label{alg:cap}
\renewcommand{\algorithmicrequire}{\textbf{Input:}}
\renewcommand{\algorithmicensure}{\textbf{Output:}}
\begin{algorithmic}[1]
\Require Gradient: $\boldsymbol{\nabla \mathcal{G}(\delta_{t})}$, Regularization parameter: $\boldsymbol{\lambda_{1},\lambda_{2}}$.

\State Initialize $\delta_0^c$, $\eta_0^c$, $U_{c,0}$, $V_{c,0}$, $z_0^c$ for all channels $c=1,\dots,C$
\For{$t=1$ to $T$}
    \For{each channel $c = 1$ to $C$}
        \State \textsc{(1) Optimistic update:} Update $U_{c,t+1}$, $V_{c,t+1}$, and $\theta_t^c$ \Comment{\cref{(1)OptimisitcUpdate}}
        \State \textsc{(2) Singular value update:} Compute $z_{t+1,1:k}^c$ \Comment{\cref{(2)Singular Values}}
        \State \textsc{(3) Perturbation update:} Compute $\delta_{t+1}^c$ \Comment{\cref{(3) low rank perturbation}}
    \EndFor
\EndFor
\Ensure $\boldsymbol{\delta_{T+1}}$
\end{algorithmic}
\end{algorithm}

\section{Experiments \& Evaluation}
\label{sec:exp}
This section details the empirical evaluation of the proposed algorithms. We first describe the experimental setup, including the baseline attack, metrics, datasets, and models used within this experimental scope. We then present and analyze the results. All code and data used in our experiments are publicly available online\footnote{\url{https://github.com/jsve96/AO-Exp-Attack}}.
\subsection{Baseline Attack}
\subsubsection*{LoRa-PGD:}
The low-rank PGD attack is a variation of the projected gradient descent (PGD) that directly searches for a low-rank structured perturbation \cite{LoRaPGD}. The $k$-th iteration of a PGD attack is the following perturbation
\[ \delta_{k}=\mathcal{P}\left(\delta_{k-1}+\epsilon \frac{\nabla_{\delta} \mathcal{L}}{||\nabla_{\delta} \mathcal{L||}}\right),\] where $\mathcal{P}$ denotes the projection on some feasible set of perturbations. For the LoRa-PGD attack, the perturbation is decomposed into two lower-rank matrices $U \in \mathbb{R}^{H \times r \times C}$ and $V \in \mathbb{R}^{r \times W \times C}$ where $r\leq \min\{H,W\}$, such that each entry of the perturbation for each channel $c \in C$ is given by 
\[ \delta_{ijc} = (U \otimes V)_{ijc} = \sum_{k=1}^{r}U_{ikc}V_{kjc}.\]
The update rule is applied independently on $U$ and $V$ such that 
\begin{equation*}
     U_{k} = U_{k-1} + \frac{\nabla_{U} \mathcal{L}}{||\nabla_{U} \mathcal{L||}} \hspace{1cm} V_{k} = V_{k-1} +\frac{\nabla_{V} \mathcal{L}}{||\nabla_{V} \mathcal{L||}}
     \end{equation*}
and the updated perturbation is $\delta_{k+1} = (U_{k+1} \otimes V_{k+1})$. Putting this together, in our universal framework, the LoRa-PGD attack has the following formulation:
\begin{equation*}
(U_{*},V_{*}) =\begin{cases}
\begin{aligned}
\argmax_{U,V} \quad & \frac{1}{B}\sum_{b=1}^{B}\mathcal{L}_{\textrm{total}}(f(x_{b}+\mathbf{\delta}),f(x_{b}))\\
\textrm{s.t.} \quad 
 & U \in \mathbb{R}^{H \times r \times C}, V \in \mathbb{R}^{r \times W \times C}\\
 & r \leq \min\{H,W\} \quad \textrm{(rank constraint)}\\
 & ||U \otimes V||_{*} \leq \tau \quad \textrm{(nuclear norm constraint)}
\end{aligned}
\end{cases}
\end{equation*}
\subsubsection*{FW-Nucl:}
The Frank-Wolfe nuclear norm group attack (FW-Nucl) obtains structured adversarial examples by constraining perturbations using nuclear group norm regularization \cite{kazemi2023minimally}. It iteratively applies the Frank-Wolfe algorithm to construct a sparse adversarial perturbation. In our universal attack formulation, FW-Nucl has the following form:
\begin{equation*}
\delta^{*}=\begin{cases}
\begin{aligned}
        \argmin_{\mathbf{\delta} \in \mathbb{R}^{H \times W \times C}} \quad & \frac{1}{B}\sum_{b=1}^{B}\mathcal{L}_{\textrm{total}}(f(x_{b}+\mathbf{\delta}),f(x_{b}))\\
        \textrm{s.t.} \quad & ||\delta||_{\mathcal{G},1,p}\leq \epsilon
\end{aligned}
\end{cases}
\end{equation*}
\subsection{Metrics}
Intersection over Union (IoU), also known as the Jaccardi Index, is a well-known measure for the similarity of two shapes or sets $A,A' \in \mathbb{R}^{n}$,
\begin{align*}
    \text{IoU}(A,A') = \frac{|A \cap A'|}{|A\cup A'|},
\end{align*}
which is often used as a loss function for bounding box regression \cite{rezatofighi2019generalized}, where $A$ denotes the predicted bounding box of the vision model and $A'$ is a ground truth location of the bounding box. Suppose there are a total of $n$ ground truth bounding boxes of the vision model using the clean frame $x_{t}$, and the adversarial example $x_{t}+\delta$ leads to $m$ predicted bounding boxes, then the IoU for frame $x_{t}$ is 
\begin{equation*}
    \textrm{IoU}_{t}=\sum_{i=1}^{n}\sum_{j=1}^{m}\text{IoU}(A_{i},A'_{j}).
\end{equation*}
We evaluate the average IoU score over all frames and report the accumulated IoU
\begin{equation*}
    \textrm{IoU}_{acc}=\frac{1}{T}\sum_{t=1}^{T}\textrm{IoU}_{t}
\end{equation*}
to assess the impact of the adversarial attack on the whole sequence. Moreover, we report the ratio of the sum of bounding boxes of the adversarial video clip and the sum of ground truth bounding boxes for the clean video clip (advBR). The Box Ratio indicates whether all ground truth bounding boxes are removed (advBR $=0$) or the object detector is fooled when additional bounding boxes appear for the adversarial video frames (advBR $>1$). Additionally, we report perceptibility, which we measure based on the mean absolute perturbation (MAP)
\begin{equation*}
    \text{MAP} = \frac{1}{H \cdot W}\sum_{i,j=1}^{H,W}\sum_{c=1}^{C}|\delta_{ij}^{c}|.
\end{equation*}
\subsection{Datasets}
\begin{table}[t]
\caption{Overview of datasets used and key attributes.}
\label{Datset Table}
\vskip 0.1in
\begin{center}
\begin{adjustbox}{width=0.99\textwidth}
\begin{tabular}{ccccc}
\toprule
    Name & Resolution & Scenes & Frames per second & Avg. frames per scene\\
\midrule
PETS 2009 S2L1 \cite{ferryman2009pets2009} & 768×576 & $7$ & 7&795\\
EPFL-RLC \cite{EPFL-RLC} & 1920×1080 &$3$& $60$ & $5000$ \\
CW4C & 1920×880& $15$ & $60$ & $7200$\\
\end{tabular}
\end{adjustbox}
\end{center}
\vskip -0.1in
\end{table}
\textbf{PETS 2009 S2L1} \cite{ferryman2009pets2009}:
The PETS 2009 dataset is a benchmark video surveillance dataset designed to evaluate algorithms for multi-camera tracking, crowd analysis, and event detection. It features synchronized footage from multiple camera views capturing various real-world scenarios, such as people walking, meeting, splitting up, or leaving objects behind. It is widely used in academic research for tasks like people tracking, group activity recognition, and anomaly detection. \begin{figure}
    \centering
    \includegraphics[width=0.9\linewidth]{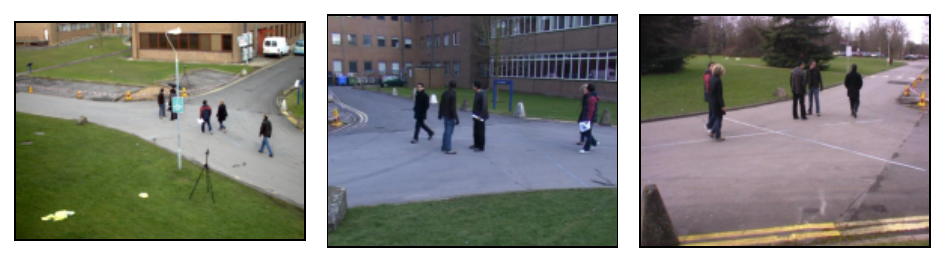}
    \caption{Shows frame number 55 of the PETS 2009 dataset for three different camera views.}
    \label{fig:Pets2009}
\end{figure}

\textbf{EPFL-RLC} \cite{EPFL-RLC}:
The EPFL-RLC dataset is a multi-camera video dataset captured at the Rolex Learning Center of EPFL using three synchronized HD cameras with overlapping fields of view. Each camera records at a resolution of 1920×1080 at 60 frames per second, with the dataset comprising 8,000 frames per view. This dataset is particularly valuable for developing and evaluating multi-view pedestrian detection and tracking algorithms \cite{chavdarova2018wildtrack}.
\begin{figure}
    \centering
    \includegraphics[width=0.9\linewidth]{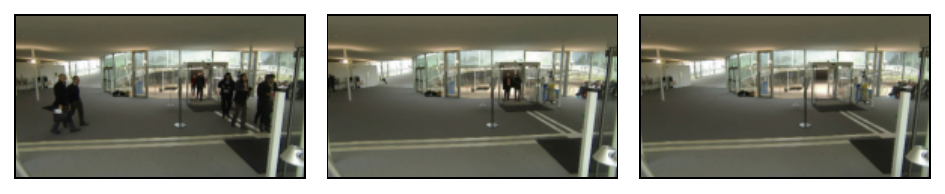}
    \caption{Shows three frames from the EPFL-RLC dataset.}
    \label{fig:placeholder}
\end{figure}

\textbf{CW4C} \footnote{ \url{https://www.coldwater.org/676/Coldwater-Area-Webcams}} (Coldwater 4 corners):
This dataset contains 15 video clips from the publicly available CW4C data \footnote{https://tinyurl.com/CW4C-Data}. The data captures 4 Corners Park, located at the intersection of Chicago St and Marshall St, in Coldwater (Michigan). In order to limit and save computational cost, we applied downsampling and modified the resolution to $960\times 440$.
\begin{figure}
    \centering
    \includegraphics[width=0.95\linewidth]{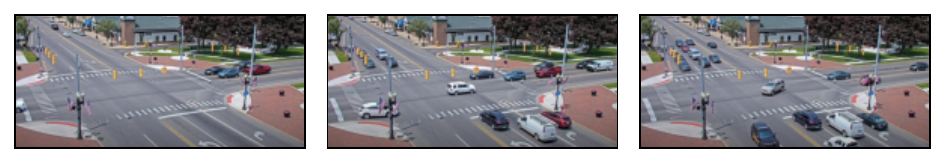}
    \caption{Shows three frames from the first video clip of the camera capturing the crossroad intersection in Coldwater (CW4C).}
    \label{fig:CW4C_frame}
\end{figure}
\subsection{Evaluation}\label{sec:Eval}
For LoRa-PGD attacks and full AO-Exp attacks, we use 100 iterations to obtain a universal perturbation. We set the regularization parameter $\lambda_{1}$ for \Cref{alg:cap} to $\lambda_{1}=0.1$ for PETS datasets, and set $\lambda_{2}=\lambda_{1}/500$ (CW4C), $\lambda_{2}=\lambda_{1}/10$ (PETS2009). We set $\lambda_1=0.75$ and $\lambda_2=0.005$ (EPFL-RLC). For FW-Nucl, we set $\epsilon=40$ for comparable results using the same perturbation budget as for AO-Exp.
\begin{figure}
    \centering
    \includegraphics[width=0.8\linewidth]{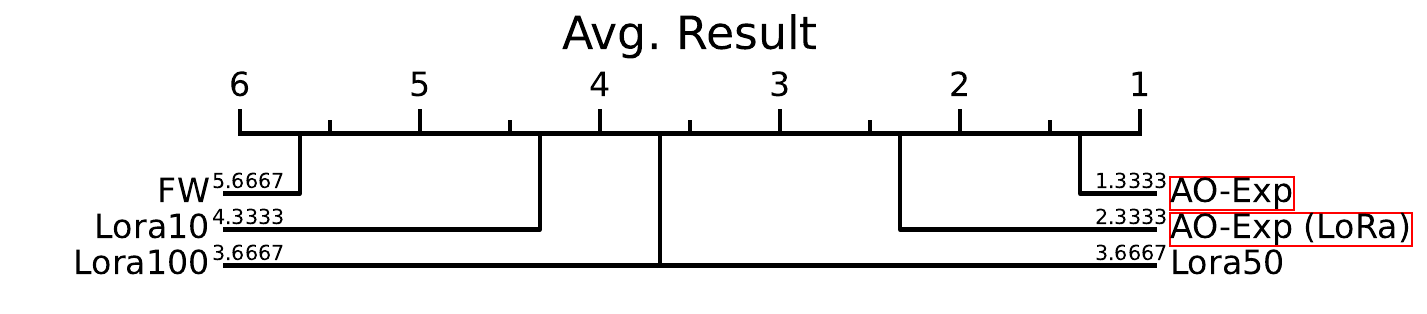}
    \caption{Shows the critical difference diagram of the average result, obtained based on the scores of $\textrm{IoU}_{acc}$, advBR, and $||\delta||_{*}$ (\Cref{ViT Table}), across all datasets for each attack method.}
    \label{fig: cd-Plot}
\end{figure}
Moreover, we set the number of iterations to 30 and use five updates for each line search. For the PGD-LoRa attacks, we report three variants with $r=10\%,50\%,100\%$ of the full rank and set the nuclear norm budget to $60$. For the low-rank adaption of AO-Exp, we only use the top value $(k=1)$ in $(\ref{(3) low rank perturbation})$ and consider 50 iterations.
\begin{table}[t]
\caption{Results of baseline attack methods and our proposed minimally structured universal attack method AO-Exp over three real-world datasets. For each method, we report the mean value and standard deviation across all scenes of the corresponding dataset. Bold values indicate best result, underlined values second best result.}
\label{ViT Table}
\begin{center}
\begin{adjustbox}{width=0.99\columnwidth}
    \begin{tabular}{clcccc}
\toprule
    Dataset &Attack Method &$ \textrm{IoU}_{acc} (\downarrow)$ &advBR $(\downarrow)$ & MAP $(\downarrow)$ &  $||\delta||_{*} \ (\downarrow)$  \\
\midrule
\multirow{6}{*}{PETS2009}     &FW-Nucl & $4.77 \pm 1.09$ &$1.04 \pm 0.25$&$\mathbf{1.2} \pm 0.3$&$\underline{36.5} \pm 5.84$ \\
  &  LoRa-PGD-10&$1.98 \pm 1.31$&$0.94\pm 0.68$&$4.0 \pm 0.4$&$39.6\pm 1.73$\\
   &LoRa-PGD-50&$1.48 \pm 1.05$&$0.67 \pm 0.41$&$4.0 \pm 0.5$&$60.9\pm 10.3$\\
   & LoRa-PGD-100&$\underline{1.22} \pm 0.91$&$0.63 \pm 0.42$&$4.0 \pm 0.3$&$60.3\pm 10.3$\\
  \cmidrule{2-6}
     &AO-Exp&$\mathbf{0.29}\pm 0.27$&$\mathbf{0.06}\pm 0.04$&$\underline{2.9}\pm 0.1$&$41.3 \pm 16.6$\\
     & AO-Exp (LoRa) & $1.88 \pm 1.38$ & $\underline{0.45} \pm 0.33$& $6.0 \pm 0.2$&$\mathbf{17.7} \pm 4.3$\\
\midrule
\multirow{6}{*}{EPFL-RLC}    &FW-Nucl &$4.83 \pm 0.96$&$0.86 \pm 0.14$& $\mathbf{5.4} \pm 2.0$&$37.54 \pm 1.53$\\
 &  LoRa-PGD-10& $0.25 \pm 0.15$&$0.31\pm 0.05$&$14.6 \pm 3.0$& $31.59 \pm 1.38$ \\
  &LoRa-PGD-50&$\mathbf{0.17} \pm 0.03$&$\underline{0.29} \pm 0.14$&$14.0 \pm 2.7$&$42.84 \pm 2.95$\\
   & LoRa-PGD-100&$\underline{0.20} \pm 0.06$&$0.37 \pm 0.11$&$14.0 \pm 3.0$&$43.5 \pm 4.3$\\
  \cmidrule{2-6}
     &AO-Exp&$0.9\pm 0.37$&$\mathbf{0.22}\pm 0.07$& $\underline{6.0} \pm 4.0$&$\underline{27.52}\pm 15.8$\\
      & AO-Exp (LoRa) &$1.39 \pm 0.60$ &$0.33 \pm 0.1$&$7.0\pm 3.0$&$\mathbf{4.7}\pm 0.93$\\
     \midrule
\multirow{6}{*}{CW4C}    &FW-Nucl &$4.64 \pm 3.69$&$0.82 \pm 0.16$&$\mathbf{1.3}\pm 0.2$&$39.4 \pm 0.5$\\
  &  LoRa-PGD-10&$2.72 \pm 2.61$&$0.48 \pm 0.25$ & $5.0 \pm 0.4$&$\underline{36.1} \pm 1.1$ \\
   &LoRa-PGD-50&$2.32 \pm 2.24$&$0.41 \pm 0.22 $& $4.0 \pm 0.2$&$65.2 \pm 6.7$\\
   & LoRa-PGD-100&$\underline{1.95}\pm 1.91$&$\underline{0.34} \pm 0.2$&$5.0 \pm 0.2$&$67.9\pm10$\\
  \cmidrule{2-6}
     &AO-Exp & $\mathbf{0.88} \pm 0.45$ &$\mathbf{0.19} \pm 0.06$&$\underline{3.8} \pm 1.0$&$37.9 \pm 8.9$\\
      & AO-Exp (LoRa) &$2.12 \pm 1.18$ &$0.42 \pm 0.17$&$5.0 \pm 4.0$&$\mathbf{14.3}\pm 7.6$\\
     \bottomrule
\end{tabular}
\end{adjustbox}
\end{center}
\end{table}

\begin{figure}
    \centering
    \begin{subfigure}[b]{0.55\linewidth}        
        \centering
        \includegraphics[width=\linewidth]{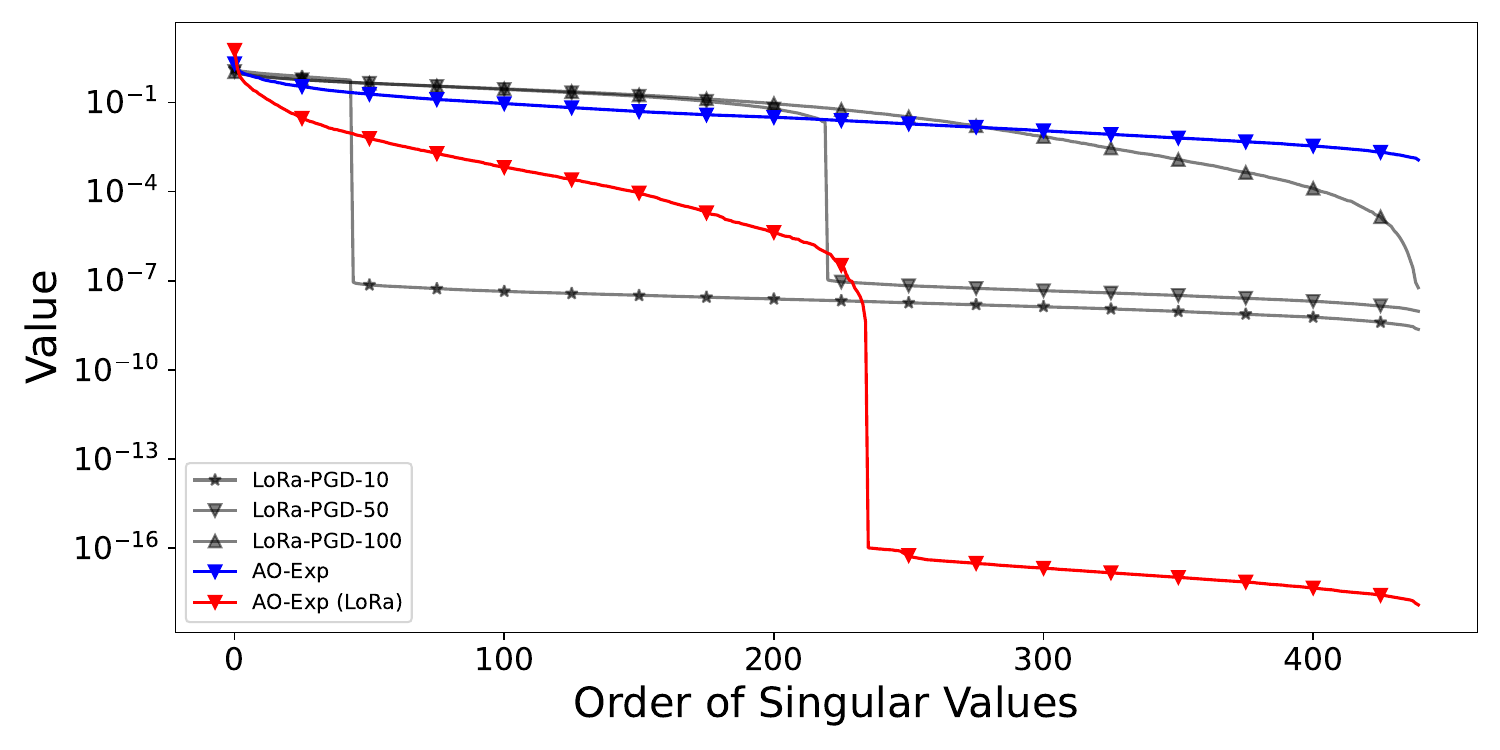}
        \caption{Median singular values of LoRa-PGD attacks, AO-Exp attack, and low-rank adaption (AO-Exp LoRa) for CW4C dataset. AO-Exp (LoRa) uses only the top singular value in \Cref{(3) low rank perturbation}.}
        \label{fig:SVD_plot_methods}
    \end{subfigure}
    \hspace{0.2cm}
    \begin{subfigure}[b]{0.35\linewidth}        
        \centering
        \includegraphics[width=\linewidth]{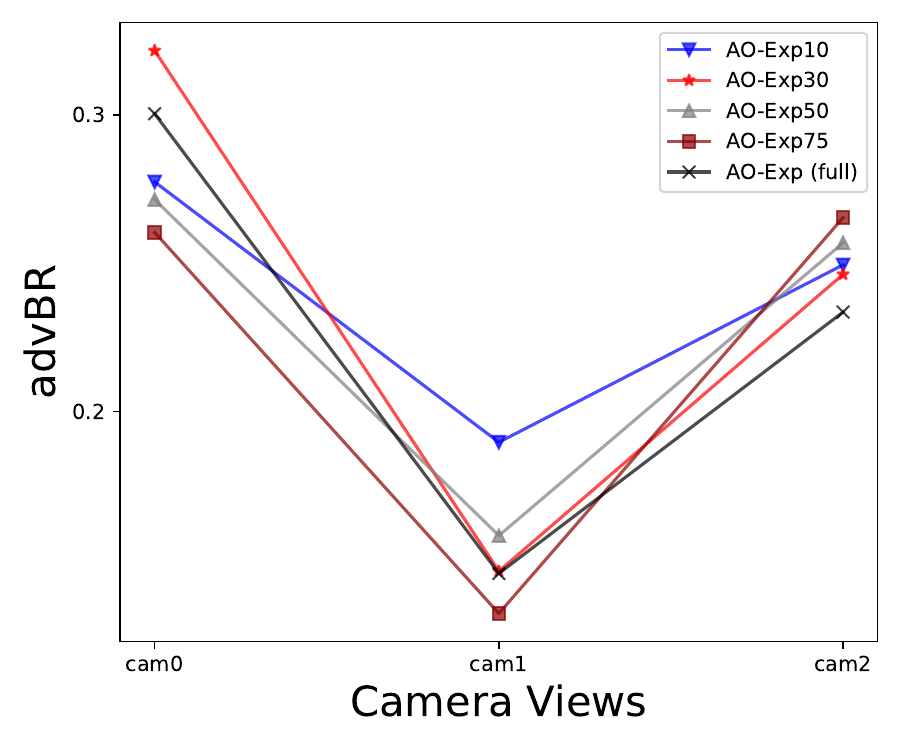}
        \caption{Shows adversarial box ratio of five variants of AO-Exp with different values of $k$ in \Cref{(3) low rank perturbation} for each camera view of the EPFL dataset.}
        \label{fig:B}
    \end{subfigure}
    \caption{Additional results for CW4C and EPFL datasets considering low-rank adaptions of AO-Exp.}
    \label{fig:roc_curve}
\end{figure}

We observe that our proposed method, AO-Exp, achieves the best adversarial box ratio across all datasets while minimizing the accumulated IoU and notably minimal nuclear norm, as shown in \Cref{ViT Table}. This also holds for the MAP, indicating stealth attacks in general. Notably, our low-rank adaption not only yields comparable results in the average adversarial box ratio but also drastically minimizes the nuclear norm of the perturbation compared to AO-Exp, as shown in \Cref{fig:SVD_plot_methods}. Our proposed attack method, AO-Exp, and its low-rank adaption surpass the considered baseline attacks across all datasets, averaged over three main metrics, see \Cref{fig: cd-Plot}. Moreover, we observe that different values of $k$ in the update rule of \Cref{alg:cap} yield similar advBR for each camera angle of the EPFL dataset, \Cref{fig:B}. Increasing the number of singular values for the construction of the structured adversarial perturbation enhances the effectiveness of the adversarial attack, as expected.

\section{Limits \& Conclusion}
\label{sec:conclusion}
In this work, we proposed a novel minimally distorted universal adversarial attack designed for video-based object detection systems. By leveraging nuclear norm regularization, our method promotes structured perturbations that primarily target the background, enabling stealthier and more natural-looking adversarial examples. To tackle the associated computational complexity, we leverage an adaptive optimistic exponentiated gradient descent approach, which improves both scalability and convergence.

Despite these promising results, our approach has some limitations. First, the current formulation assumes a static camera setup, limiting its applicability to dynamic camera scenarios. Second, the attack's performance is sensitive to the choice of hyperparameters, such as the nuclear norm weight and Frobenius regularization, which may require task-specific tuning.

Future work may explore extending this approach to dynamic camera settings, extending the work to object tracking, developing adaptive or learned hyperparameter strategies, and integrating semantic or temporal consistency constraints to improve generalizability and stealth in more complex real-world scenarios. Additionally, one may explore countermeasures and defenses tailored specifically to structured and temporally consistent adversarial attacks.
\subsubsection{\ackname} 
This research was funded by the German Federal Ministry of Labour and Social Affairs through the establishment of a Junior Research Group on Artificial Intelligence at the Federal Institute of Occupational Safety and Health (BAuA). The
presented results contribute to the development and evaluation of reliable and safe AI for industrial applications, with the overarching aim of laying the scientific foundations necessary to meet the requirements of the European Machinery Directive (2023) and the European AI Act (2024).

\bibliographystyle{splncs04}
\bibliography{047-main}

\end{document}